\documentclass[letterpaper,twocolumn,10pt]{article}

\usepackage{usenix,epsfig} 
\usepackage{xspace}
\usepackage[inline, shortlabels]{enumitem}
\usepackage{booktabs}
\usepackage{cprotect}
\usepackage{amsmath}
\usepackage{array}
\usepackage[table,svgnames]{xcolor} 
\usepackage{pifont} 

\usepackage{graphicx}
\usepackage{subcaption}
\usepackage{nicefrac}
\usepackage{amsmath}

\usepackage{algorithm}
\usepackage{algpseudocode}
\definecolor{commentgray}{HTML}{696969}

\newcommand{\algrule}[1][.2pt]{\par\vskip.3\baselineskip\hrule height #1\par\vskip.5\baselineskip}

\newcommand{\cmark}{\ding{51}} 
\newcommand{\xmark}{} 

\newcommand{\engine}{{\sffamily LARRY}\xspace}
\newcommand{\loadbal}{{\sffamily SAL}\xspace}

\newcommand{\trail}{{\sffamily TRAIL}+\xspace}
\newcommand{\maxconcurr}{{\sffamily FCFS}\xspace}
\newcommand{\nopreempt}{{\sffamily No-Preempt}\xspace}
\newcommand{\ptc}{{\sffamily P2C}\xspace}
\newcommand{\random}{{\sffamily Random}\xspace}

\newcommand{\todo}[1]{{\color{red}TODO: #1}}
\newcommand{\ferdi}[1]{{\color{blue}Ferdi: #1}}

\newcommand{\blfootnote}[1]{%
  \begingroup
  \renewcommand\thefootnote{}\footnote{#1}%
  \addtocounter{footnote}{-1}%
  \endgroup
}

\makeatletter
\patchcmd{\maketitle}
	{\@maketitle}
	{\@maketitle\vspace{-5em}}
	{}
	{}
\makeatother

\begin{document}

\date{}

\title{\vspace{-5em}Is the GPU Half-Empty or Half-Full? Practical Scheduling Techniques for LLMs}

\author{
{\rm Ferdi Kossmann$^{1,*}$, Bruce Fontaine$^2$, Daya Khudia$^2$, Michael Cafarella$^1$, Samuel Madden$^1$}\\
{$^1$MIT CSAIL, $^2$Databricks}\\
{\{ferdik, michjc, madden\}@csail.mit.edu, \{bruce.fontaine, daya.khudia\}@databricks.com}
}

\maketitle

\thispagestyle{empty}

\subsection*{Abstract}

Serving systems for Large Language Models (LLMs) improve throughput by processing several requests concurrently. However, multiplexing hardware resources between concurrent requests involves non-trivial scheduling decisions. 
Practical serving systems typically implement these decisions at two levels: First, a \emph{load balancer} routes requests to different servers which each hold a replica of the LLM. Then, on each server, an \emph{engine-level scheduler} decides when to run a request, or when to queue or preempt it. Improved scheduling policies may benefit a wide range of LLM deployments, and can often be implemented as ``drop-in replacements'' to a system's current policy. 
In this work, we survey scheduling techniques from the literature and from practical serving systems.
We find that schedulers from the literature often achieve good performance but introduce significant complexity. In contrast, schedulers in practical deployments often leave easy performance gains on the table but are easy to implement, deploy and configure.
This finding motivates us to introduce two new scheduling techniques, which are both easy to implement, and outperform current techniques on production workload traces.


\section{Introduction}
\label{sec:introduction}

Large Language Models (LLMs) are capable of powerful text generations that power chat bots, code assistants, and more.\blfootnote{* Work done while at Databricks.}
For a given request, the LLM predicts the next response token by conditioning on previous tokens in the sequence. This requires the LLM to project the previous tokens to \emph{key} and \emph{value} vectors, which are used by the attention mechanism. LLM serving systems avoid repeatedly recomputing these projections by storing them in a \emph{KV cache}. The KV cache resides in GPU memory and grows as a request's token sequence grows and the prediction context becomes larger. A single request's KV cache can occupy significant chunks of GPU memory that are only freed once the request finishes. For example, the KV cache of Llama-3 70B occupies 2.7 GB for a single sequence of 8192 tokens.

LLM serving systems increase throughput by propagating several requests through the LLM as one \emph{batch}. However, due to the memory consumption of each request's KV cache, the GPU's memory capacity limits how many requests may be processed concurrently. Since the response length of the LLM is not known a priori, a request's memory consumption is also not known a priori.
State-of-the-art serving systems employ Paged Attention~\cite{vllm} to dynamically allocate more memory to a request as its KV cache grows. With Paged Attention, the serving system avoids over-allocating memory to requests with short sequence lengths. This allows the serving system to improve throughput by running more requests concurrently.

\begin{figure}[t]
    \centering
    \includegraphics[width=0.43\textwidth]{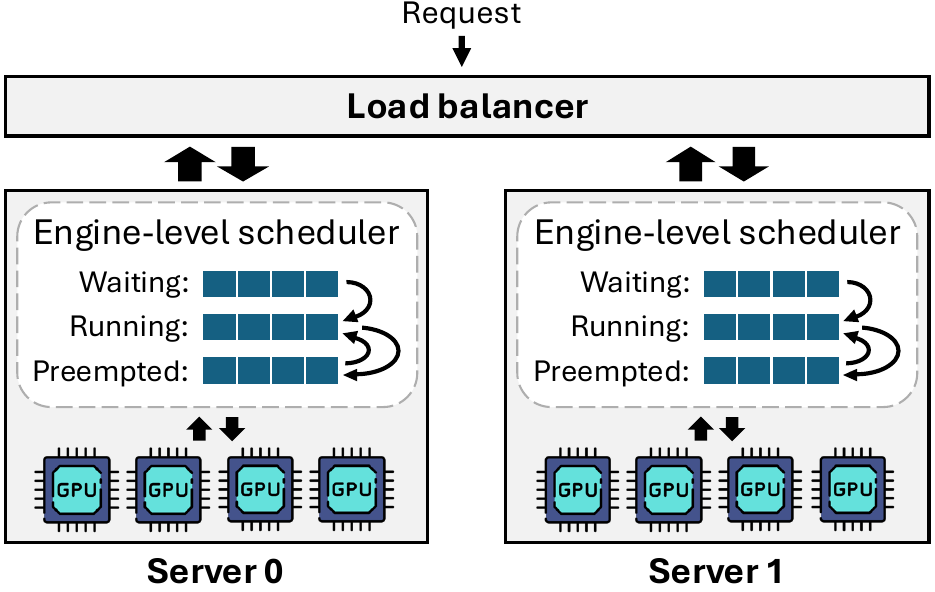}
    \caption{Typical serving architecture where request scheduling occurs at two levels.}
    \label{fig:sched-levels}
    \vspace{-1.5em}
\end{figure}

However, the serving system may also run out of memory once the requests' growing KV caches exceed the GPU's memory capacity. In this case, the serving system must \emph{preempt} one of the running requests, evict its KV cache and pause its execution. Once the system schedules the preempted request for resumption, restoring the KV cache is an expensive operation, often implemented by recomputing the KV cache from scratch~\cite{flexgen, vllm}. 
To avoid preemptions, serving systems typically control how many requests run in parallel and may hold requests in a \textit{waiting queue} before dispatching them for execution. 

To achieve low queuing delays and low preemption rates, LLM serving systems must implement a scheduling policy that determines which requests to dispatch for execution, and how to batch requests. Typically, scheduling decisions are made at two levels (Figure~\ref{fig:sched-levels}). First, a request hits a load balancer that decides which server a request should be routed to. Here, each server  holds a replica of the LLM which allows to horizontally scale the overall throughput of the system (\emph{data parallelism}).
Second, each server runs a serving engine that decides which requests to run at what time. In Figure~\ref{fig:sched-levels}, engine-level scheduling decisions are depicted as shuffling requests between a waiting queue, a running queue, and a preempted queue.

Given how commonly this architecture occurs in LLM serving systems, improved scheduling policies are applicable to many systems and can improve the operational cost and quality of service of many LLM deployments. 
Nevertheless, the scheduling of LLM requests remains underexplored: There are relatively few research works that address the LLM scheduling problem, and many require significant changes to practical serving systems (Sections~\ref{sec:engine} and~\ref{sec:load_bal}). 
As a consequence, practical systems today implement a potpurri of different scheduling techniques. For example, vLLM~\cite{vllm}, TensorRT-LLM~\cite{trt-llm}, and SGLang~\cite{sglang} implement a combined total of six different scheduling policies. 

\begin{table} 
    \centering
    \caption{Overview of surveyed scheduling techniques}
    \vspace{-0.5em}
    \small
    \begin{tabular}{@{\hskip 5pt}l@{\hskip 14pt}@{\hskip 13pt}c@{\hskip 10pt}@{\hskip 7pt}c@{\hskip 9pt}}
     \toprule
        & \textbf{Drop-in}                     & \textbf{Designed} \\ 
        & \textbf{replacement\footnote{.}}     & \textbf{for LLMs\footnote{.}}  \\
        \midrule
        \textbf{Engine-level schedulers}       &        &        \\
        \hspace{1em}FCFS                       & \cmark & \xmark \\
        \hspace{1em}No-Preempt                 & \cmark & \xmark \\
        \hspace{1em}S$^3$~\cite{s3}            & \xmark & \cmark \\
        \hspace{1em}PiA~\cite{pia}             & \xmark & \cmark \\
        \hspace{1em}FastServe~\cite{fastserve} & \xmark & \cmark \\
        \hspace{1em}TRAIL~\cite{trail}         & \xmark & \cmark \\
        \hspace{1em}LTR~\cite{ltr}             & \xmark & \cmark \\

        \hspace{1em}\engine \textit{(New)}     & \cmark & \cmark \\ 
        \midrule
        \textbf{Load balancers}                &        &        \\
        \hspace{1em}RR                         & \cmark & \xmark \\
        \hspace{1em}Random                     & \cmark & \xmark \\
        \hspace{1em}P2C                        & \cmark & \xmark \\
        \hspace{1em}Llumnix~\cite{llumnix}     & \xmark & \cmark \\
        \hspace{1em}\loadbal \textit{(New)}    & \cmark & \cmark \\
        \bottomrule
        \end{tabular}
    \label{tab:techniques}
    \vspace{-1em}
\end{table}

\blfootnote{\hspace{-0.5em}$^1$Doesn't require changes outside of the scheduler nor upfront training of additional prediction models.}\blfootnote{\hspace{-0.5em}$^2$Account for properties that are specific to LLM requests, opposed to schedulers that apply general-purpose policies to LLM serving.}In this work, we give a comprehensive overview of LLM scheduling techniques. We first survey the techniques that are implemented in practical systems and the techniques that are proposed in the literature. We compare some of these techniques by implementing them inside the same representative serving system. Through this, we find that schedulers from the literature often outperform practical schedulers but introduce significant complexity. Some of them require profound changes to the serving system, which go beyond just modifying the scheduler. Others require the user to train additional, application-specific prediction models. In contrast, more practical systems rely on scheduling techniques that are easy to implement, deploy, and configure.

To bridge the gap between theoretical research and practical applications, we propose two techniques that are both easy to implement and deploy and that also enhance system performance. First, we introduce \engine\footnote{\textbf{L}oad-\textbf{A}daptive \textbf{R}equest \textbf{R}eordering for Low Latenc\textbf{y}}, an engine-level scheduler that reorders requests in the queue based on (i) the expected resource consumption of the requests and (ii) the current system load. The logic for this reordering uses metrics that are readily available within the scheduler, and we implemented \engine in just 20 lines of code in vLLM. Similarly, we present \loadbal\footnote{\textbf{S}erver-\textbf{A}ware \textbf{L}oad Balancer}, a token-aware load balancer that requires only basic metrics on requests and server load. We implemented \loadbal in 30 lines of code in the system described in Section~\ref{sec:method}.

\engine and \loadbal can be implemented as ``drop-in replacements'' to the policies inside many current serving systems. Specifically, they neither require modifications outside of the scheduler, nor do they need to train additional prediction models. Nevertheless, we find that \engine and \loadbal outperform current techniques and therefore present a practical opportunity for easy performance gains.


Table~\ref{tab:techniques} summarizes the techniques surveyed in this paper. We describe each technique in more detail in Sections~\ref{sec:engine},~\ref{sec:load_bal}, and~\ref{sec:ours}. In summary, our contributions are as follows.

\begin{itemize}
    \item We review current scheduling policies, both from the literature and from practical LLM serving systems.
    \item We compare various scheduling techniques by implementing them inside the same, representative system.
    \item We introduce an engine-level scheduler and a load balancer. The techniques can easily be implemented inside many LLM serving systems and outperform current techniques on production workload traces. 
\end{itemize}

\section{Engine-Level Schedulers}
\label{sec:engine}




An LLM serving engine multiplexes hardware resources between inference requests. This gives rise to a scheduling problem, which we describe in Section~\ref{sec:engine-problem}.

In this paper, we assume that scheduling techniques may leverage optimizations that are commonly found in state-of-the-art serving engines.
Specifically, we assume the serving engine supports Continuous Batching~\cite{orca}, where requests may start and stop running at any generation iteration. We also assume that the engine supports Paged Attention~\cite{vllm}, which allows it to dynamically allocate non-contiguous chunks of memory to a request's growing KV cache. Finally, we assume that the serving engine supports Chunked Prefill~\cite{sarathi}, where requests in the prefill phase may be batched together with requests in the decode phase.
Many popular serving engines support these optimizations, including vLLM~\cite{vllm}, SGLang~\cite{sglang}, TensorRT-LLM~\cite{trt-llm}, DeepSpeed~\cite{deepspeed}, and more.

\subsection{Problem Definition}
\label{sec:engine-problem}


%
%
%

\begin{figure*}[t]
    \centering
    \begin{minipage}{0.23\textwidth}
        \centering
        \includegraphics[width=\textwidth]{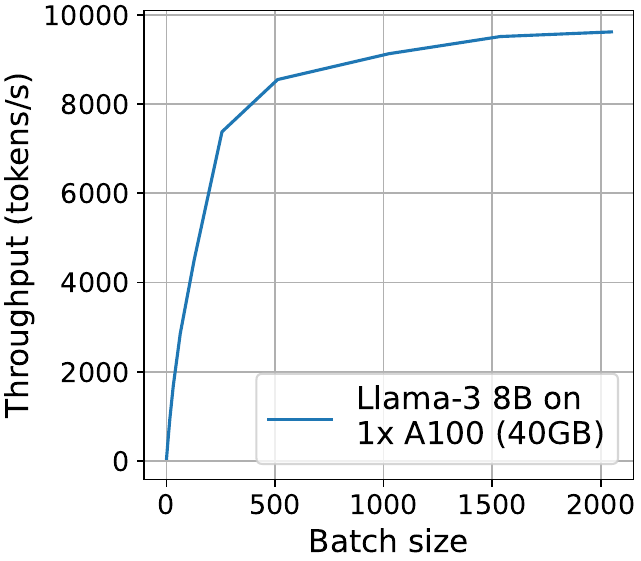}
        \caption{Throughput during the decode phase as a function of the batch size.}
        \label{fig:throughput}
    \end{minipage}%
    \hfill
    \begin{minipage}{0.23\textwidth}
        \centering
        \includegraphics[width=\textwidth]{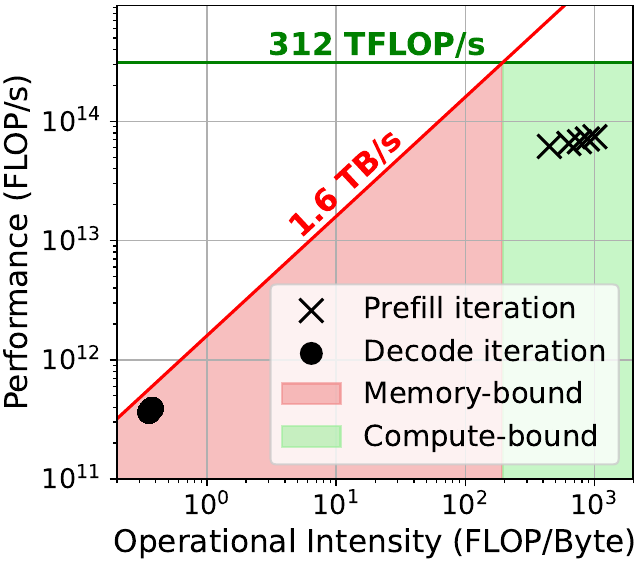}
        \caption{Roofline analysis of SW-Chat requests for Llama-3 8B on an A100 GPU~\cite{splitwise}.}
        \label{fig:roofline}
    \end{minipage}%
    \hfill
    \begin{minipage}{0.23\textwidth}
        \centering
        \includegraphics[width=\textwidth]{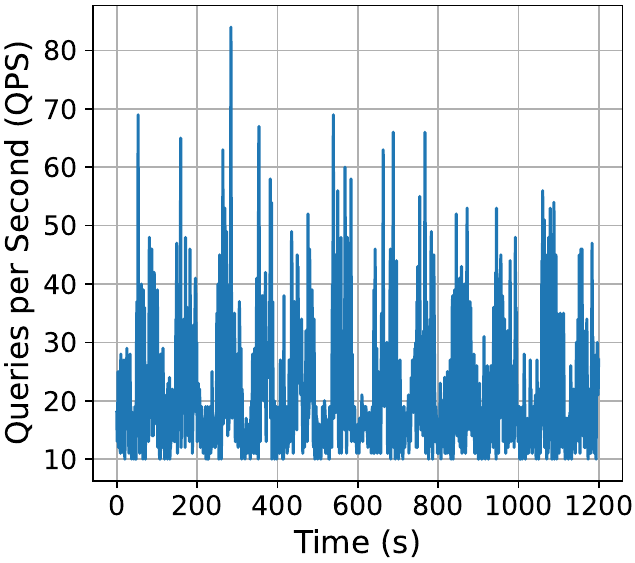}
        \caption{20-minute excerpt from a representative production request trace~\cite{workload}.}
        \label{fig:workload}
    \end{minipage}%
    \hfill
    \begin{minipage}{0.23\textwidth}
        \centering
        \includegraphics[width=\textwidth]{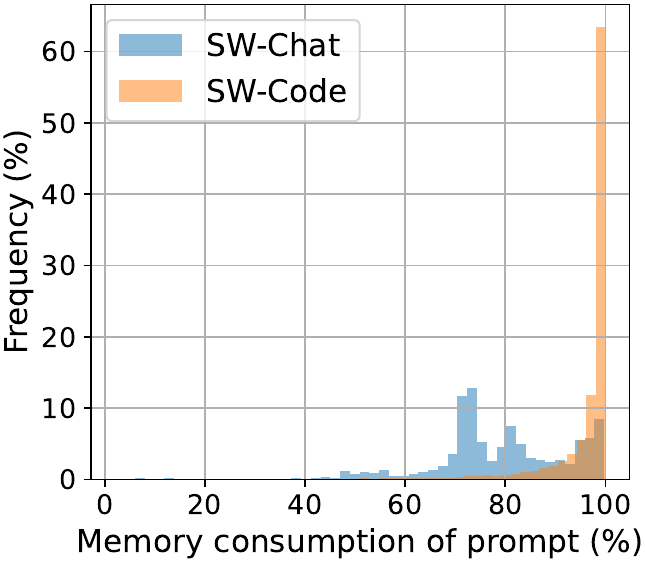}
        \caption{Fraction of a request's memory consumption used by prompt~\cite{splitwise}.}
        \label{fig:ratios}
    \end{minipage}%
    \vspace{-0.5em}
\end{figure*}

A serving engine may propagate several requests through the LLM together as one \emph{batch}. Figure~\ref{fig:throughput} shows how the batch size affects throughput for a Llama-3 8B model on a an NVIDIA A100 GPU (40GB memory). At low batch sizes, the GPU spends more time loading data from memory to the arithmetic untis than on arithmetic itself (i.e., the forward passes are \emph{memory-bound}). The loaded data is dominated by the 16GB model weights. As the batch size increases, these model weights only need to be loaded once and can be used to compute the next token of several requests, therefore improving throughput. However, including more requests in a forward pass also increases the amount of arithmetic and the throughput gains diminish for large batches, where more time is spent on arithmetic as on loading data (i.e., the forward passes become \emph{compute-bound}).

A serving engine should always use large batch sizes in order to maximize throughput.
However, the GPU's memory capacity limits the number of requests that can run concurrently. This limitation arises because each LLM request produces intermediary state called the \emph{KV cache}, which may occupy significant fractions of GPU memory. For example, the KV cache of a sequnce with 8192 tokens for Llama-3 70B occupies 2.7GB of memory. 
The size of the KV cache depends on the number of tokens that the LLM bases its prediction on (i.e., the size of the context window). Therefore, the memory requirements vary between requests and depend on the request's sequence length. 

State-of-the-art engines \emph{dynamically} allocate more memory to a request as the request's sequence length and memory requirement grows~\cite{vllm}. 
At first, such engines may dispatch many requests to run concurrently but this might cause the GPU to run out of memory later as the requests' KV caches grow. In such a scenario, some requests must be \emph{preempted}, which involves evicting their KV cache to free up memory and allocate it to other requests.
If the engine suspects that dispatching a request will lead to a preemption, it may delay executing it and instead hold the request in a \emph{waiting queue}. In summary, an engine's scheduling policy consists of the following elements. 

\vspace{-1em}\paragraph{Dispatching.} An engine must decide when to dispatch a request for execution. This decision must be made for all requests that are waiting for initial dispatching, and all requests that have been preempted and are waiting to be rescheduled. In its decision, the engine may consider the current memory consumption of the running requests, as well as properties of the running and queued requests.

\vspace{-1em}\paragraph{Preemption.} The engine must decide when and which requests to preempt. The engine may consider how much memory the system still has available, as well as the cost of preempting a request, both in terms of quality of service and the incurred overhead of evicting its KV cache.

\vspace{-1em}\paragraph{KV cache eviction.} When a request is preempted, its KV cache must be evicted from GPU memory. The most common approach of coing so is to delete the KV cache and recompute it as the request is rescheduled. Alternatively, the KV cache may also be transfered to CPU memory and which allows reloading it into GPU memory as the request is rescheduled. However, due to low bandwidth between GPU and CPU, this can be slower than recomputing the KV cache~\cite{vllm}. In this work, we therefore assume that the KV cache of a preempted request is recomputed.

\subsection{Current Approaches}

\paragraph{First-Come-First-Served (\maxconcurr).} Waiting requests are dispatched First-Come-First-Served (FCFS) and the user defines a limit on how many requests may run concurrently. Setting a high concurreny limit generally improves throughput but may lead to more preemptions. Setting a low concurrency limit may underutilize the GPU resources and lead to low throughput. This policy is implemented in engines including vLLM~\cite{vllm}, TensorRT-LLM~\cite{trt-llm}, and SGLang~\cite{sglang}.


\vspace{-1em}\paragraph{FCFS with no preemptions (\nopreempt).} A request's memory consumption has an upper bound defined by the smaller of either, (i) the maximum context length of the LLM,
or (ii) the prompt length and a maximum response length defined by the user.
The request's KV cache may reach this maximum size but the request may also finish before reaching it (e.g., because the LLM predicts an End-Of-Sequence token). 
This policy allocates the maximum memory to each request and only dispatches a request when enough memory can be allocated for it. As a result, no request is preempted, but some requests may not require all of their allocated memory. Requests in the waiting queue are dispatched in FCFS order. This policy is implemented in engines including TensorRT-LLM~\cite{trt-llm} and ORCA~\cite{orca}.

\vspace{-1em}\paragraph{S$^3$~\cite{s3}.} S$^3$ improves the throughput of systems that don't implement Paged Attention and need to allocate the maximum size the KV cache may reach for each request. To avoid always allocating memory for the worst case, S$^3$ uses a Distil-BERT predictor~\cite{distilbert} that predicts how much memory a request will occupy and only allocates memory according to this prediction. 
To handle inaccurate predictions, S$^3$ implements a supervisor that handles the case where the predicted memory requirement is smaller than the actual requirement. The supervisor aborts requests that exceed the estimated memory requirement and defragments the memory. 

Since we focus on systems that implement Paged Attention, we don't include S$^3$ in our evaluation. Notably, S$^3$ has been proposed before Paged Attention. The supervisor in S$^3$ cannot be implemented as a drop-in replacement to schedulers in current systems, as it also requires separate changes to how memory is managed~\cite{vllm, trt-llm, deepspeed}.




\vspace{-1em}\paragraph{PiA~\cite{pia}.} PiA enhances systems that do not implement Continuous Batching. When batching requests together, such systems can only return a request's response once the last request in a batch finished execution. Furthermore, new requests can only be dispatched once the responses of the previous batch are completely generated. PiA uses the LLM to estimate each request's response length. PiA then forms batches with requests that are expected to have similar response lengths.

State-of-the-art serving systems implement Continuous Batching~\cite{vllm, deepspeed, sglang, vllm, orca}, including the serving system used in our experiments. Since PiA specifically addresses short comings that don't exist in these systems, we don't include PiA in our evaluation. Notably, PiA was proposed before Continuous Batching~\cite{orca}.

\vspace{-1em}\paragraph{TRAIL~\cite{trail}.} TRAIL proposes a predictor that estimates how many tokens of a request's response are left to be generated. TRAIL leverages this estimation to implement a Shortest Predicted Remaining Processing Time
(SPRPT) policy~\cite{sprpt}. Specifically, TRAIL prioritizes requests who have few tokens left to be generated (as estimated by the predictor).
A running request may be preempted in favour of a newly-arrived request, if the new request is estimated to have fewer remaining tokens than the running request. To avoid unnecessary preemption overheads, TRAIL defines a parameter $0\leq c\leq 1$ and only allows requests to be preempted in the beginning, when $\nicefrac{num\_generated\_tokens}{est\_response\_length} < c$. 

TRAIL is not a drop-in replacement to current schedulers as it uses the layer activations to estimate the remaining response length. In many practical implementations, the scheduler does not have access to the layer activations~\cite{vllm, trt-llm, deepspeed}. 

In our evaluation, we implement \trail which is an upper bound to TRAIL's performance. \trail has access to the ground-truth response lengths, effectively providing TRAIL with perfect predictions at no runtime overhead.

\vspace{-1em}\paragraph{LTR~\cite{ltr}.} The Learning-To-Rank (LTR) scheduler takes a similar approach to TRAIL but, instead of predicting the \emph{absolute} output length of a request, LTR just \emph{ranks} requests according to their output length. While this is an easier prediction task, it still allows LTR to first dispatch requests that have few tokens left to be generated. 

LTR can be implemented inside the scheduler of many practical serving systems. However, LTR requires to train a ranking model in an offline phase.
Furthermore, computing resources (i.e., GPU cycles and memory) need to be multiplexed between the ranking model and the served LLM.


\vspace{-1em}\paragraph{FastServe~\cite{fastserve}.} FastServe proposes a Multi-Level Feedback Queue where requests are held in different queues that denote different priorities. FastServe cleverly shuffles requests between the queues and re-assesses a request's priority after it has run for a given amount of time (a \textit{quantum}).
Given that FastServe re-assesses a request's priority while it is running but before it has finished, requests may frequently be paused and resumed.
FastServe makes this approach feasible by proactively moving KV caches between host memory and GPU memory, such that a request can be resumed without needing to wait for the KV cache to be loaded.

FastServe's proactive KV cache management requires additional implementation effort that goes beyond simply replacing the scheduling policy of current serving engines. For example, FastServe cannot be implemented as a drop-in replacement to scheduling policies in vLLM, SGLang, TensorRT-LLM, or ORCA.

\section{Load Balancing}
\label{sec:load_bal}

A serving system may replicate the LLM onto several servers. This allows the system to increase throughput by splitting work between the replicas (data parallelism). However, the serving system must balance the load evenly between servers to effectively scale throughput. This is achieved through a \emph{load balancer}, which we discuss in this section.

\subsection{Problem Definition}


If a serving system uses several server instances, a request first hits a load balancer which aims to route requests to servers such that the load is equally split amongst them. Balancing the load between servers is not trivial since different inference requests require different amounts of resources. 
Specifically, variations in  resource consumption occur because of two factors, which we discuss below.

\vspace{-1em}\paragraph{Request sequence length.} 
When generating tokens, LLMs store intermediate state in a KV cache. The memory consumption of the KV cache linearly depends on the number of tokens that the LLM bases its prediction on (i.e., the size of the context window). Therefore, a request's sequence length determines (i) how much memory the system needs to allocate to a request, and (ii) how long the requests runs and for how long the memory needs to be allocated. 
In many practical deployments, the sequence length may vary widely between requests~\cite{burstgpt, splitwise}. Furthermore, the load balancer doesn't know a request's response length a priori, and therefore doesn't know its ultimate resource consumption a priori.

\vspace{-1em}\paragraph{Processing phase of request.} LLM inference happens in two distinct phases. During the \emph{prefill phase}, the LLM processes the user prompt and populates the KV cache with vector representations of the prompt tokens. The LLM may batch several tokens of the user prompt together and propagate them through the LLM as one batch. This typically makes the prefill phase \emph{compute-bound}, meaning that the GPU spends more time on arithmetic than on moving data from memory to the arithmetic units.
Upon completion of the prefill phase, the LLM produces the first response token and the request enters the \emph{decode phase}. 
During the decode phase, each forward pass only propagates a single token through the LLM for a given request. This typically makes the decode phase \emph{memory-bound}, meaning that more time is spent moving the model weights and KV cache from memory to the arithmetic units, than is spent on the arithmetic itself.

Figure~\ref{fig:roofline} visualizes the difference between these two phases using the roofline model. We separately propagate five requests through Llama-3 8B on an A100 GPU. The crosses represent prefill iterations, the dots decode iterations. Figure~\ref{fig:roofline} shows how the prefill iterations incur higher operational intensity and are compute-bound, whereas the decode iterations are memory-bound. 


\subsection{Current Approaches}

Many serving systems rely on general-purpose orchestration platforms to deploy model servers. For instance, numerous serving systems~\cite{kserve, mlflow, clipper} are built on Kubernetes~\cite{kubernetes}, often in combination with additional services, 
such as KNative~\cite{knative}, Istio~\cite{istio}, or Envoy~\cite{envoy}. These underlying platforms allow serving systems to implement their own, application-specific load balancers, but also provide general-purpose, LLM-agnostic load balancers. In practice, some serving systems, such as KServe~\cite{kserve} or MLFlow~\cite{mlflow}, continue to rely on the general-purpose load balancers of the orchestration platform. As a result, this paper not only focuses on LLM-specific load balancers but also considers the general-purpose load balancers of popular platforms that serving systems are built upon.

\vspace{-1em}\paragraph{Round-Robin (RR).} RR simply assigns requests to servers in an incremental fashion, where request $i$ is routed to server $i\mod{\;num\_servers}$. This policy is implemented in Istio~\cite{istio-lb} and Envoy~\cite{envoy-lb}. KNative implements a variant of this~\cite{knative-lb}.

\vspace{-1em}\paragraph{Random.} This policy samples a server uniformly at random. The policy is implemented in Istio~\cite{istio-lb} and Envoy~\cite{envoy-lb}.

\vspace{-1em}\paragraph{Power of Two Random Choices (\ptc).} \ptc samples two servers uniformly at random and then chooses the server with fewer incoming TPC connections (i.e., fewer in-flight requests). \ptc is implemented in Envoy~\cite{envoy-lb} and KNative~\cite{knative}.


\vspace{-1em}\paragraph{Llumnix~\cite{llumnix}.} Llumnix is an LLM serving system that can \textit{reschedule} requests to other server instances while the request is running. 
Llumnix proposes a mechanism to migrate a request's KV cache from one server to another while only incurring a minimal interruption to the running request's progress.
Llumnix leverages this mechanism in a clever scheduling policy to route and migrate requests between server instances while improving load balancing, performance isolation, and resource fragmentation.

The Llumnix mechanism for KV cache migration is not implemented in many LLM serving systems and requires modifying both, the load balancer and the serving engine.

\section{Proposed Techniques: \engine and \loadbal}
\label{sec:ours}

The scheduling techniques surveyed in Sections~\ref{sec:engine} and~\ref{sec:load_bal} either require additional modifications to components separate to the scheduler (i.e., are not ``drop-in replacements"), need to train and run additional models, or are direct adoptions of general-purpose, LLM-agnostic scheduling techniques.
This inspires us to propose two additional scheduling techniques.
These techniques are easy to implement, deploy and configure and can be used as a drop-in replacements for current scheduling techniques. However, they still explicitly account for the decisive properties of LLM requests.

\subsection{Engine-Level Scheduler: \engine}

In practice, many applications see unpredictable, short-term fluctuations in the received Queries-Per-Second (QPS).
For example, Figure~\ref{fig:workload} visualizes an excerpt of a production QPS trace~\cite{workload} that is noted to be representative for model serving~\cite{infaas, cascadeserve, shepherd}.
As the QPS varies, so does the pressure on GPU memory:
Under high QPS, many queries are competing for memory; reserving a large chunk of memory may prevent many other queries from being dispatched.
In contrast, under low QPS, competition for memory is lower and reserving a large chunk of memory delays fewer other queries (or no queries at all if memory is abundant).
We now describe \engine\footnote{\textbf{L}oad-\textbf{A}daptive \textbf{R}equest \textbf{R}e-ordering for Low Latenc\textbf{y}} which prioritizes queries according to their anticipated memory demand \emph{and} the current system load.

Similar to schedulers in practical systems~\cite{vllm, sglang, deepspeed, trt-llm}, \engine estimates a query's memory consumption solely based on the prompt length. Figure~\ref{fig:ratios} visualizes how much of a request's total KV cache was used to store tokens from the prompt. The requests were recorded from a chat bot (SW-Chat) and a code copilot (SW-Code) application on Azure OpenAI~\cite{splitwise}. In both workloads, over 70\% of the tokens stored in most KV caches are tokens from the prompt. Similar trends can be observed in further request traces~\cite{burstgpt}. Given that (i) the prompt dominates a request's memory consumption in many applications, and (ii) the KV cache only grows slowly after the prefill phase, \engine dispatches a request once the available memory can hold its prefill KV cache.


\engine deprioritizes requests with large memory consumption (i.e., large prompts) when there is high pressure on memory (i.e., a long waiting queue).
Rather, \engine aims to schedule large requests once memory pressure decreases and obtaining large chunks of memory doesn't block many other queries in the waiting queue.
However, to avoid starvation of large queries, \engine also prioritizes queries that have been waiting in the queue for a longer time.
Specifically, \engine assigns a score to each query in the waiting queue.
\engine then sorts the waiting queue according to the requests' scores and dispatches requests in decreasing order of their scores. \engine stops dispatching requests when either (i) all requests are dispatched, or (ii) a request is reached that cannot be scheduled anymore (due to insufficient memory or the current batch already reaching the maximum token size).
The score $score(r)$ for request $r$ is given by Equation~\ref{eq:engine}.

\begin{equation}
\label{eq:engine}
    score(r) = \alpha * wait\_time(r) - queue\_len*memory(r)
\end{equation}

Parameter $\alpha$ weighs how much waiting time in the queue should be prioritized over a request's memory consumption. In practice, choosing a high value for $\alpha$ leads to lower tail latencies (e.g., p99 latencies) while sacrificing latency closer to the median (e.g., p50 latencies). We further evaluate this in Section~\ref{sec:eval_single_server}.

\subsection{Load Balancer: \loadbal}

A request may need to wait in a waiting queue before being dispatched because either (i) the next batch to be propagated through the LLM is already filled with the maximum number of tokens, or (ii) the engine-level scheduler won't dispatch the request because there is not enough memory available.

\loadbal\footnote{\textbf{S}erver-\textbf{A}ware \textbf{L}oad Balancer} quantifies the load each server $s$ after adding request $r$, both in terms of the queued prefill tokens and the currently available memory. \loadbal then routes request $r$ to server $s$ with the lowest load.  \loadbal quantifies the load $load(s, r)$ as stated in Equation~\ref{eq:lb}.

\begin{multline}
\label{eq:lb}
    load(s, r) 
    = \max\left( \beta * (memory(r) - free\_mem(s)), 
    \right. 
    \\
    \left. \nicefrac{queued\_tokens(s, r)}{max\_tokens\_per\_batch} 
    \right)
\end{multline}

Where $\beta = \nicefrac{(\mu_{in} + \mu_{out})}{\mu_{out}}$ is the approximate, average rate at which memory is freed up because requests finish. $\mu_{in}$ is the average number of input tokens and $\mu_{out}$ is the average number of output tokens --- these averages can easily be recorded during online serving. For example, $\beta_{SW-Chat} = \nicefrac{1365}{211}$ and $\beta_{SW-Code} = \nicefrac{2074}{27}$. $memory(r)$ is the memory required for request $r$'s input tokens, and $queued\_tokens(s, r)$ is the number of queued tokens on server $s$ plus the number of prefill tokens of $r$.

Equation~\ref{eq:lb} captures how long a request is estimated to wait in the waiting queue (either until enough memory is available, or until all queued prefill tokens have been processed).
\loadbal periodically polls the number of queued tokens and the amount of available memory from the servers. Between polling intervals, \loadbal keeps track of these statistics by assuming no request finished in the meantime. I.e., if \loadbal routes a request with $N$ tokens to a server, it updates adds $N$ to $queued\_tokens$ and the required memory for $N$ tokens to $memory$. We find that polling the server for these statistics incurs negligible overhead, which allows for frequent polling. In our implementation, we poll 10 times per second.





\section{Methodology}
\label{sec:method}

We implemented all evaluated scheduling policies inside a system based on vLLM~\cite{vllm} and evaluated the policies on two production workload traces and report common performance metrics. In the following, we describe our benchmarking methodology in more detail.

\paragraph{Workloads.}
We evaluate scheduling policies by emulating workload traces that are recorded from production settings. We emulate the incoming Queries-Per-Second (QPS) by issuing queries against the system according to the recorded time stamps of a trace of Azure Function calls~\cite{workload}. This workload trace has been noted to be representative of model serving loads in several works~\cite{infaas, cascadeserve, shepherd}. We linearly scale the recorded QPS such that the system is neither under-provisioned nor over-provisioned, and also investigate the quality of service when varying the workload factors. Figure~\ref{fig:workload} shows an excerpt of the recorded workload trace.

\begin{figure}
    \centering
    \includegraphics[width=\linewidth]{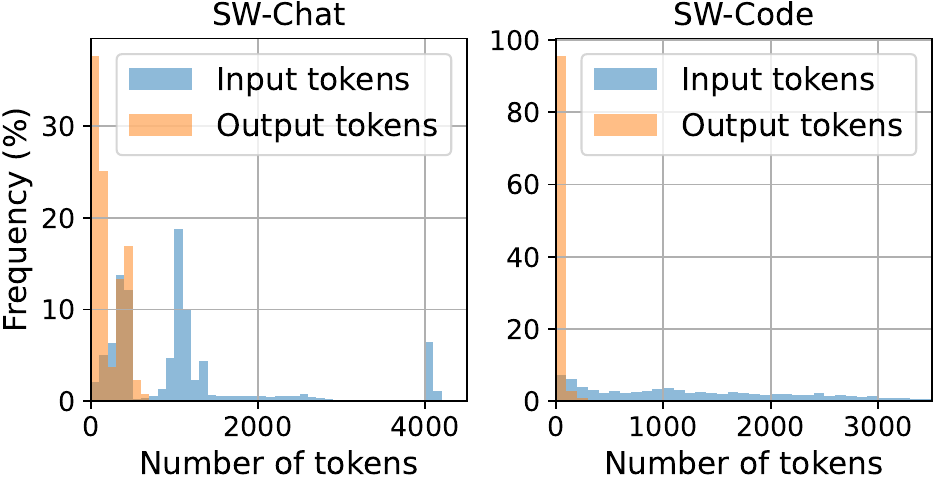}
    \caption{Distribution of input and output tokens for SW-Chat and SW-Code workloads~\cite{splitwise}.}
    \label{fig:in_out}
    \vspace{-1em}
\end{figure}

We emulate the input and output tokens of requests through traces recorded in production by Azure OpenAI~\cite{splitwise}. The trace only records the token length of the input prompt and the generated output, but not the precise input prompt. We emulate the requests by issuing requests against the system that consist of a prompt with random tokens but the same token length as the recorded prompt in the trace. We then enforce that the system produces a response that contains the exact same output length as in the recorded trace. We evaluate the schedulers on two input/output traces. One is recorded from a chat application and one from a coding co-pilot. Figure~\ref{fig:in_out} displays the distribution of the input and output lengths for both applications.

\vspace{-0.25em}
\paragraph{Implementation.} We implement the engine-level schedulers inside vLLM\footnote{Commit 57792ed469956c7e580f982d590fcacea882570b (08/22/2024)}~\cite{vllm}. Currently, vLLM uses a single driver process that receives queries and marshalls them into the scheduler's waiting queue. The data marshalling consists of several steps such as tokenization, and the driver process is also used for further tasks such as running the scheduler itself. As a consequence, the driver process becomes oversubscribed at high QPS and fails to keep up with adding queries to the waiting queue. This creates a back pressure, where queries are not queued in the waiting queue but are instead waiting to be added to the waiting queue, leaving the scheduler unaware of their existence. 

This back pressure prevents us from evaluating how different schedulers perform at high QPS, since the oversubscribed driver process effectively limits the rate at which queries are added to the waiting queue and therefore, the rate at which the scheduler receives queries. To evaluate scheduling techniques at high QPS, we therefore modify vLLM and mitigate the data marshalling overheads to occur outside the driver process (e.g., the tokenization happens in a separate process before the query is issued to the system). These modifications allow queries to be added to the waiting queue at high rates. We verified for each experiment, that the rate at which queries are added to the waiting queue precisely reflects the rate at which queries are issued according to the workload trace.

When evaluating multi-server deployments (\S\ref{sec:eval_multiple_servers}), we host each vLLM server as a separate Ray~\cite{ray} actor. We use one additional process which runs the load balancer and makes routing decisions for all queries as they arrive.

\vspace{-0.25em}
\paragraph{Evaluation Metrics.} We evaluate the scheduling techniques according to five metrics that quantify performance and user experience.

\begin{itemize}
    \item \textbf{Time To First Token (TTFT).} Time elapsed from issuing the request until the first token is generated. This includes the time the request spent in the waiting queue and completing the prefill phase.
    \item \textbf{Normalized TTFT.} TTFT divided by the number of input tokens. This captures that users may expect requests with short inputs (e.g., short prompts typed into a chat box) to return faster than requests with long inputs (e.g., summarizing a large file). \cite{tumanov-metrics} discusses the drawbacks of unnormalized TTFT in more depth.
    \item \textbf{Total generation time (TGT).} Total time elapsed from the request being issued until the request's output is completely generated.
    \item \textbf{Serving capacity.} We issue requests at time intervals dictated by production workload trace~\cite{workload}, that is noted to be representative for model serving~\cite{infaas, cascadeserve, shepherd}. We evaluate the quality of service when linearly scaling the Queries-Per-Second (QPS) of that workload trace. Since we observe low preemption rates ($<0.1\%$) for all schedulers and scaling factors, we focus on how sensitive the TTFT is to scaling the workload when using different scheduling techniques. Specifically, we focus on the p50 TTFT and p95 TTFT.
\end{itemize}

\paragraph{Hardware.} We evaluate the scheduling techniques on two hardware setups. First, we evaluate on servers with H100 GPUs and an Intel Xeon Platinum 8468V CPU with 48 cores. Second, we evaluate on servers with A100 GPUs, with 40GB memory and an AMD EPYC 7J13 CPU with 64 cores. 

\section{Performance on a Single Server}
\label{sec:eval_single_server}

\begin{figure*}[t]
    \vspace{-1.5em}
    \begin{minipage}{\textwidth}
        \includegraphics[width=\linewidth]{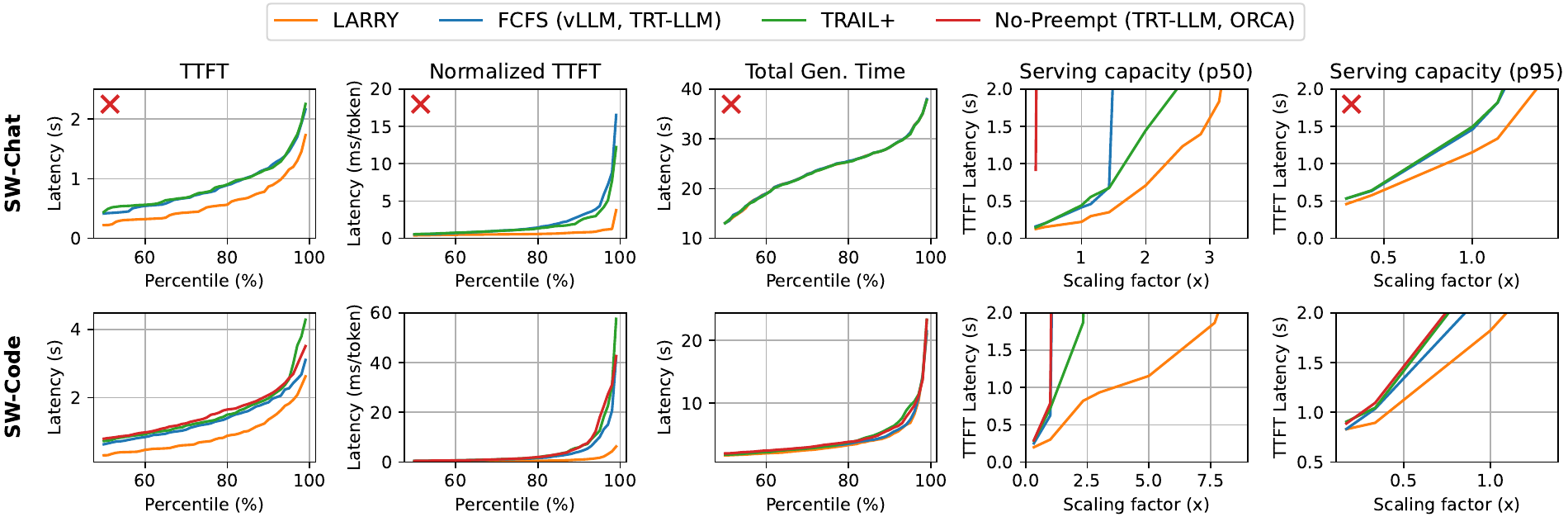}
        \caption{Scheduling techniques on a single server with two H100 GPUs running Llama-3 70B (FP8 quantized weights).}
        \label{fig:h100_single}
    \end{minipage}
    \begin{minipage}{\textwidth}
        \vspace{1em}
        \includegraphics[width=\linewidth]{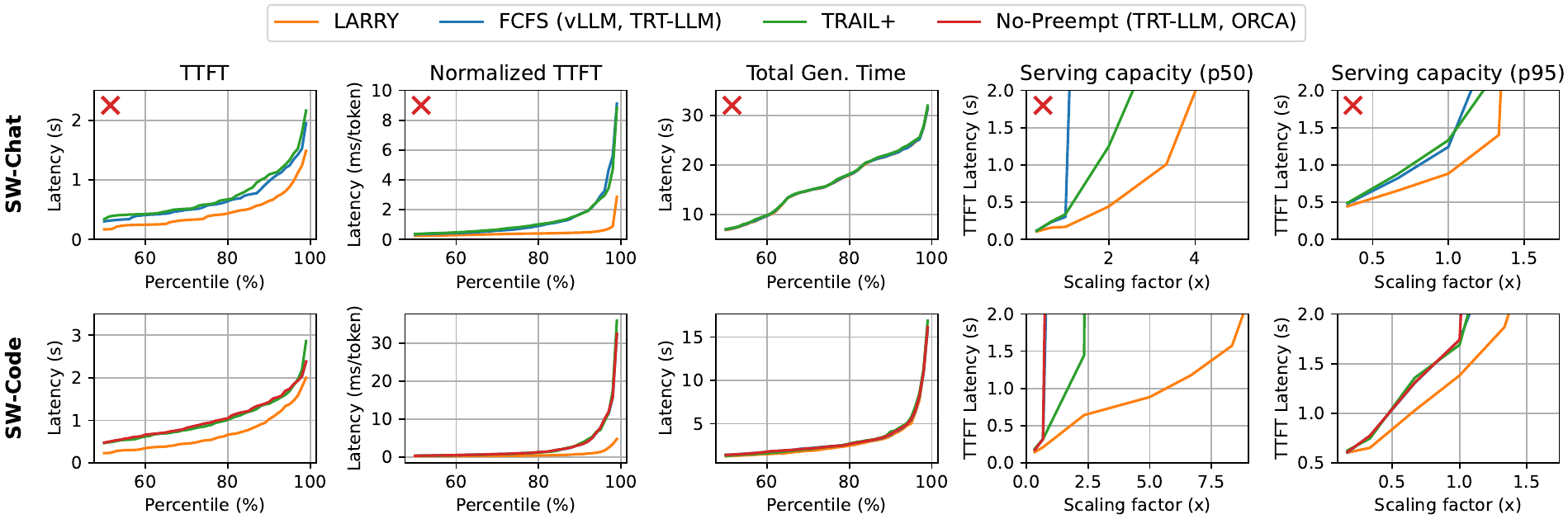}
        \caption{Scheduling techniques on a single server with an A100 GPU (40GB memory) running Llama-3 8B.}
        \label{fig:a100_single}
    \end{minipage}
    \vspace{-0.7em}
\end{figure*}

We first evaluate the engine-level schedulers in isolation and only consider single-server deployments where load balancing becomes obsolete. First, we evaluate the engine-level schedulers for Llama-3 70B (with FP8 quantized weights) on a server with two H100 GPUs (2-way tensor parallelism). Next, we evaluate the schedulers for Llama-3 8B on a server with one A100 GPU (40GB memory). We plot our findings in Figures~\ref{fig:h100_single} and~\ref{fig:a100_single} respectively. 

\vspace{-1em}\paragraph{Evaluated techniques.} We implement all scheduling techniques inside the same, representative serving engine (\S\ref{sec:method}). We focus on methods that are compatible with the optimizations implemented in state-of-the-art serving systems, specifically with Paged Attention, Chunked Prefill, and Continuous Batching. We therefore don't consider PiA~\cite{pia} and S$^3$~\cite{s3} in our evaluation. We implement the popular FCFS strategy~\cite{trt-llm, sglang,vllm,deepspeed} as well as \nopreempt~\cite{trt-llm, orca} (\S\ref{sec:engine}). The remaining methods aim to approximate the Shortest-Remaining-Processing-Time (SRPT) policy~\cite{fastserve, ltr, trail}. TRAIL is the most recent of these works and proposes a predictor to estimate the number of tokens that remain to be generated~\cite{trail}. We implement \trail, which gives TRAIL access to the ground-truth response lengths at virtually no runtime overhead. In contrast to TRAIL, LTR suggests a predictor that \emph{ranks} requests according to response length, instead of predicting the \emph{absolute} number of remaining tokens~\cite{ltr}. Since \trail has access to the ground-truth response lengths, the hardness of TRAIL's prediction task is inconsequential. Finally, like TRAIL, FastServe frequently preempts requests to approximate SRPT more accurately~\cite{fastserve}. To avoid predicting response lengths, FastServe approximates SRPT through a Multi-Level Feedback Queue. \trail avoids the need for such approximations sine it has access to the ground-truth response lengths.

\begin{figure}
    \centering
    \includegraphics[width=\linewidth]{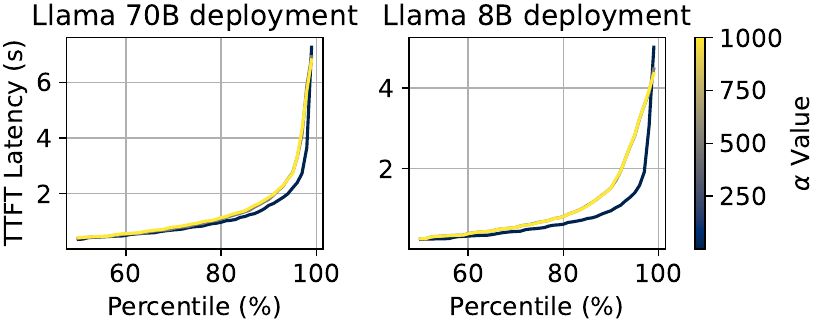}
    \caption{TTFT latencies of \engine for different values of $\alpha$. The ``Llama 70B deployment'' uses Llama-3 70B (with FP8 quantized weights) on two H100 GPUs. The ``Llama 8B deployment'' uses Llama-3 8B on one A100 GPU (40GB). Both plots run the SW-Chat workload~\cite{splitwise}.}
    \label{fig:alpha}
    \vspace{-1em}
\end{figure}

\begin{figure*}[t]
    \begin{minipage}{\textwidth}
        \includegraphics[width=\linewidth]{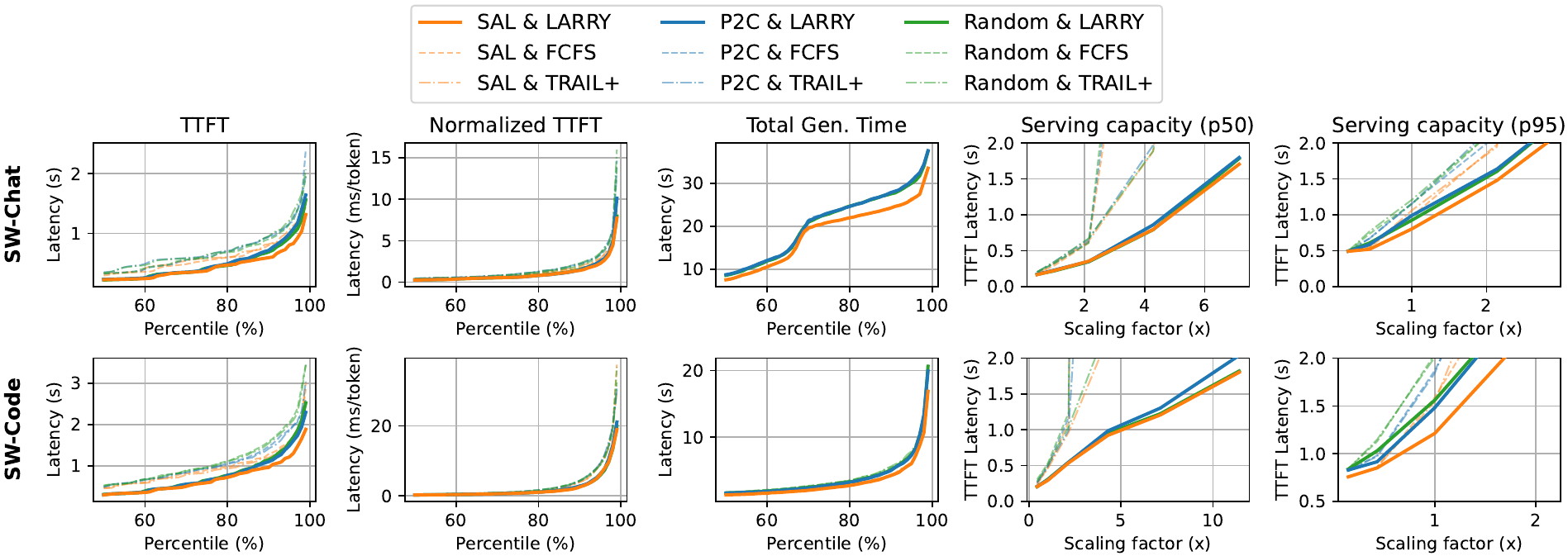}
        \caption{Scheduling techniques on four servers with two H100 GPUs, each running Llama-3 70B (FP8 quantized weights).}
        \label{fig:h100_multi}
    \end{minipage}
    \begin{minipage}{\textwidth}
        \vspace{1em}
        \includegraphics[width=\linewidth]{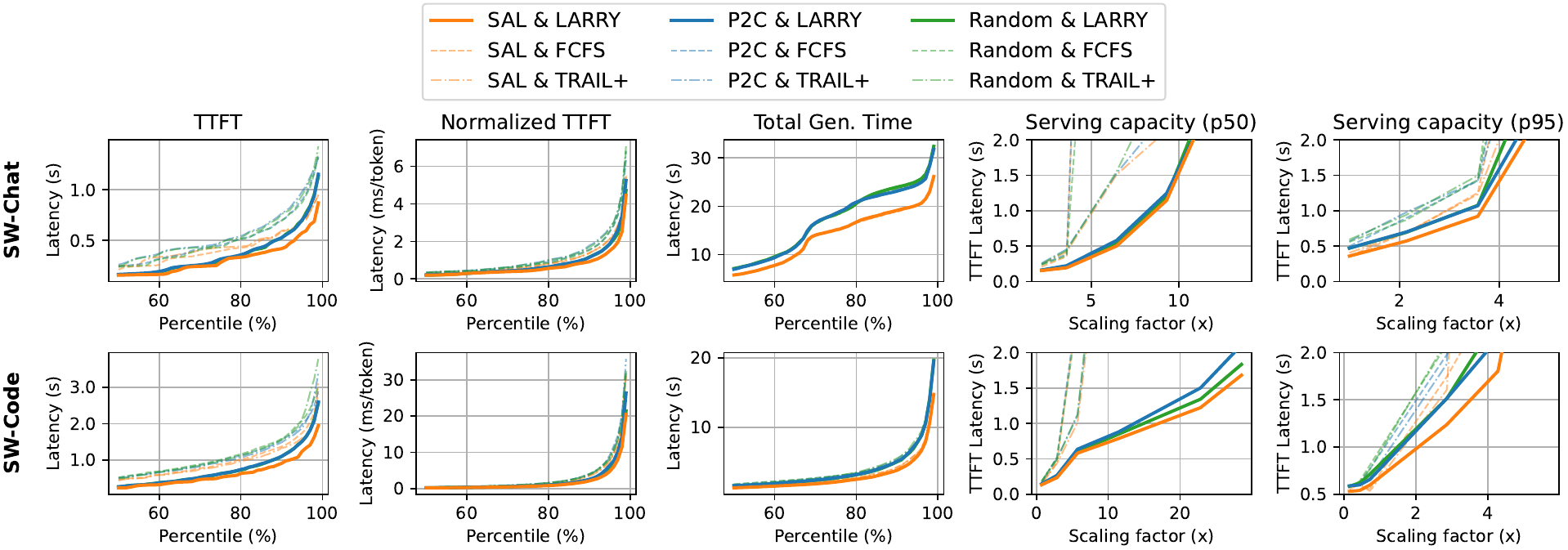}
        \caption{Scheduling techniques on eight servers with an A100 GPU (40GB memory), each running Llama-3 8B.}
        \label{fig:a100_multi}
    \end{minipage}
\end{figure*}

In Figures~\ref{fig:h100_single},~\ref{fig:a100_single},~\ref{fig:h100_multi}, and~\ref{fig:a100_multi}, we set $\alpha = 1$ for \engine. Figure~\ref{fig:alpha} shows \engine's performance for different values of $\alpha$.
We ran \trail with several choices for $c$. On our workloads, \trail achieves the best performance with $c=0$ and we therefore set $c$ to 0 in all our experiments. $c=0$ avoids frequent preemptions which is desirable for workloads with relatively short response lengths. Figure~\ref{fig:in_out} shows that both, SW-Chat and SW-Code, have shorter responses than prompts on average. 

\vspace{-1em}\paragraph{Time-To-First-Token (TTFT).} By avoiding Head-Of-Line (HOL) blocking, \engine achieves lower queueing times which results in a lower TTFT. Specifically, \engine's p50 TTFT is $1.8\times-2.1\times$ lower than the one of the next-best method and its p95 TTFT is $1.2\times-1.4\times$ lower than the one of the next-best method. \engine's Normalized p50 latency is $1.3\times-1.5\times$ lower than the one of the next-best method, and its Normalized p95 TTFT is $3.3\times-5.5\times$ lower than the one of the next-best method. \trail and \maxconcurr perform similar in terms of TTFT. On some deployments, \trail incurred starvation of large requests which lead to high tail latencies. Notably, \trail is designed for workloads with short input lengths and long output lengths, which significantly differs from the application traces used in our experiments.
The TTFT latencies of \nopreempt were so high on SW-Chat, that we didn't plot them (beyond plot limits). Since SW-Code is dominated by prefills, processing fewer requests concurrently has a smaller impact on performance. As a result, \nopreempt did significantly better on SW-Code than SW-Chat.

\vspace{-1em}\paragraph{Total Generation Time (TGT).} The TGT is generally dominated by the decode phase. For all techniques, we saw low preemption rates of less than $0.1\%$. As a result, all methods achieve similar TGT latencies. SW-Code has slightly more variation than SW-Chat since it has shorter response lengths (i.e., differences in TTFT show more significantly). \nopreempt's TGT on SW-Chat is beyond plot limits.

\vspace{-1em}\paragraph{Serving capacity.} 
We now evaluate how the quality of service changes as the workload's QPS is linearly scaled up by different factors (\S\ref{sec:method}). Compared to other methods, \engine was least sensitive to scaling the QPS up. This is reflected both, in the p50 TTFT and p95 TTFT plotted in Figures~\ref{fig:h100_single} and~\ref{fig:a100_single}. \engine's performance benefits are more pronounced when considering the p50 TTFT, since \engine prioritizes requests with short prompts under heavy load. This allows \engine to achieve high throughput and short waiting times for most requests, which is reflected in a low p50 TTFT. However, \engine's performance benefits are smaller when considering the p95 TTFT, which emphasizes the waiting times of larger requests that are deprioritized by \engine.  
Similarly, \trail also achieves low p50 TTFTs at higher scaling factors. Like \engine, \trail achieves this effect by avoiding HOL blocking. Specifically, in our workloads, the number of output tokens and input tokens is slightly correlated. This is also found to be true in other workload traces~\cite{burstgpt}. Therefore, \trail approximately prioritizes requests with smaller prefills and therefore (vaguely) prioritzes requests that may be dispatched earlier. However, since \trail doesn't prioritize requests directly according to their prefill size, it faces higher degrees of HOL blocking than \engine. Furthermore, since \trail doesn't account for starvation, its performance benefits are not as pronounced when considering p95 latencies. \nopreempt is not shown in some of the plots as its TTFT was beyond the plot limits for all scaling factors.

\vspace{-1em}\paragraph{Impact of $\alpha$ on \engine.} Figure~\ref{fig:alpha} shows the TTFT latency of \engine for different values of $\alpha$. For both subfigures, $\alpha\in\{1, 500, 1000\}$. Generally, low values of $\alpha$ (e.g., $\alpha=1$) perform better for most latency percentiles. Low $\alpha$ values allow \engine to avoid Head-Of-Line (HOL) blocking as requests with low memory requirements are prioritized over requests with high memory requirements, when memory pressure is high (\S\ref{sec:ours}). In contrast, for high values of $\alpha$ (e.g., $\alpha=1000$), one request with high memory requirement may block many others that may have lower memory requirements and could be dispatched immediately. In fact, for very high values of $\alpha$ (larger than 1000), \engine degenerates to First-Come-First-Served (FCFS) --- this happens when $\alpha$ is so high that the waiting time dominates each request's score in Equation~\ref{eq:engine}. However, Figure~\ref{fig:alpha} also shows how high values of $\alpha$ reduce tail latencies in \engine's TTFT (e.g., the p99 TTFT latency). Tail latencies in TTFT mainly capture requests with large prefills~\cite{tumanov-metrics}. For low values of $\alpha$, such requests get deprioritized in favour of requests with shorter prefills. As $\alpha$ is set to higher values, large requests are deprioritized less aggressively. However, if $\alpha$ is set to very high values and \engine degenerates to FCFS, the tail latency may increase again, if the system struggles under frequent HOL blocking and throughput becomes a problem. This is the reason that \engine is able to outperform FCFS for all latency percentiles in Figures~\ref{fig:h100_single},~\ref{fig:a100_single},~\ref{fig:h100_multi}, and~\ref{fig:a100_multi}.
In summary, we recommend to choose low values for $\alpha$ (e.g., $\alpha=1$), which make \engine most effective at avoiding HOL blocking. While such settings will lead to the largest performance improvements for most percentiles, applications that are sensitive to tail latencies (e.g., p99 TTFT) should choose higher values for $\alpha$.

\section{Performance on Multiple Servers}
\label{sec:eval_multiple_servers}


We now evaluate the scheduling policies when using several server instances. Each server runs its own replica of the model and we evaluate combinations of load balancers and engine-level schedulers. 
Figure~\ref{fig:h100_multi} shows the performance of different techniques on a deployment of four servers, each containing two H100 GPUs and running a Llama-3 70B replica, whose weights are FP8 quantized, with 2-way tensor-parallelism. Similarly, Figure~\ref{fig:a100_multi} shows the performance of different scheduling techniques on a deployment with eight servers, each containing one A100 GPU (40GB) and running a Llama-3 8B replica. 

While Section~\ref{sec:eval_single_server} focuses on evaluating different engine-level schedulers, this section focuses on the load balancers. In our evaluation, all load balancers achieve the best performance in combination with \engine. For all combinations of scheduling techniques, we observe low preemption rates of <0.1\%. In general, in combination with the same engine-level scheduler, \loadbal improves or matches the performance of other load balancers. We now describe our findings in more detail. 

\vspace{-1em}\paragraph{Evaluated techniques.} We focus our evaluation on load balancers that are compatible with standard implementations of data parallelism, where servers aren't required to communicate amongst each other. We therefore do not evaluate Llumnix~\cite{llumnix}, which is a load balancer that requires the transfer of KV caches between servers. We evaluate \ptc, as it is a commonly used load balancer that measures and considers the current load of a server~\cite{envoy, knative}. We further evaluate \random, which is also widely used and achieves similar performance as policies like Round-Robin, while being more robust to antagonistic patterns in the workload~\cite{envoy, istio}. We exclude \nopreempt as an engine-level scheduler due to its poor performance in Section~\ref{sec:eval_single_server} --- we verified that \nopreempt does not achieve competitive performance in a multi-server setting either.

\vspace{-1em}\paragraph{Time-To-First-Token (TTFT).} We find that the load balancer has a smaller effect on queueing delays than the engine-level schedulers (\S\ref{sec:eval_single_server}). Among the benchmarked load balancers, \loadbal generally achieves the lowest TTFT latencies. Compared to the next-best technique and when comparing all load balancers in combination with \engine, \loadbal achieves $1.0\times-1.5\times$ lower p50 TTFT, and $1.2\times-1.3\times$ lower p95 TTFT. Furthermore, \loadbal achieves $1.0\times-1.2\times$ and $1.0\times-1.2\times$ lower Normalized TTFT. Overall, \ptc and \random achieve similar performance.

\vspace{-1em}\paragraph{Total Generation Time (TGT).} Unlike the engine-level schedulers (\S\ref{sec:eval_single_server}), the load balancers have a significant effect on the TGT. The TGT is generally dominated by the decode phase. In our experiments, the load balancer impacts the inter-token latency and therefore has an effect on the decode latency and TGT. Specifically, by distributing tokens more evenly between servers, there are fewer batches that have disproportionately many tokens. In our experiments, we enforce that no batch contains more than 1024 tokens (i.e., no forward pass becomes heavily compute-bound). However, batches with several 100s of tokens do incur a significantly larger runtime than batches with few tokens. Therefore, the inter-token latency is smaller for batches with fewer tokens than batches with many tokens. 
Evenly balancing the batch size decreases the number of overly full batches and, hence, improves the inter-token latency and TGT. While \ptc and \random achieve similar TGT latencies, \loadbal's token-aware routing strategy improves the TGT. Specifically, \loadbal decreases the p50 TGT by $1.1\times - 1.3\times$ and the p95 TGT by $1.1\times - 1.3\times$ over the next-best method.

\vspace{-1em}\paragraph{Serving capacity.} As for TTFT, we find that the engine-level scheduling can have a a larger impact on serving capacity than the load balancer. However, we can also observe differences between load balancers. On SW-Code, all load balancers achieve a similar p50 TTFT for different workload scaling factors. Furthermore, \random and \loadbal achieve a similar p50 TTFT for all scaling factors evaluated on the H100 GPU. For the remaining evaluations, \loadbal improves on the other techniques while there neither \ptc nor \random is able to outperform the other.

\section{Related Work}

We review load balancers and engine-level schedulers for general LLM applications in Sections~\ref{sec:load_bal} and~\ref{sec:load_bal}. Further works have suggested LLM request schedulers for different scenarios. SGLang~\cite{sglang} proposes a scheduling technique for workloads where requests share long prefix sequences. In such scenarios, the prefill can be cached and shared between requests. The goal of the proposed scheduler is to achieve high hit rates of cached prefills. In another work, S-LORA~\cite{slora1, slora2} proposes scheduling techniques when serving a large collection of Low-Rank Adapters (LoRA)~\cite{lora}.

Similar to \engine, many prior works have proposed scheduling techniques that adapt to both, the nature of requests and the system load~\cite{vetl,brad,infaas,dynamoe}. Similar to \loadbal, several works estimate the resource requirements of a query to make scheduling decisions~\cite{stage, alizadeh}. However, these works study significantly different applications and their techniques do not apply to LLM serving.


\section{Limitations}

In our evaluation, we ensure that requests are added to the request queue with little overhead. Specifically, the rate at which requests are added to the waiting queue reflects the Queries-Per-Second (QPS) recorded in the workload trace. If a serving system incurs significant overheads when adding requests to the waiting queue, our general findings about the schedulers still hold true, but their performance numbers might differ from the ones reported in Sections~\ref{sec:eval_single_server} and~\ref{sec:eval_multiple_servers}. We discuss this further in Section~\ref{sec:method}.

This work studies the \emph{online} scheduling problem. This stands in contrast to \emph{offline}, throughput-oriented applications, where all requests arrive at once and the system tries to process all of them in as little time as possible. We leave the evaluation of offline scenarios for future work.

\section{Conclusion}

In this work, we evaluated several scheduling techniques and compared their performance when implementing them inside the same, representative serving system. We propose two additional methods, a load balancer (\loadbal) and an engine-level scheduler (\engine) which are easy to implement, deploy and configure, while improving performance over existing techniques. Our evaluation is based on production workload traces and emulates a chat application and a code copilot~\cite{splitwise}.

{\footnotesize \bibliographystyle{acm}
\bibliography{references}}

\begin{thebibliography}{10}

\bibitem{tumanov-metrics}
{\sc Agrawal, A., Agarwal, A., Kedia, N., Mohan, J., Kundu, S., Kwatra, N., Ramjee, R., and Tumanov, A.}
\newblock Etalon: Holistic performance evaluation framework for llm inference systems, 2024.

\bibitem{sarathi}
{\sc Agrawal, A., Kedia, N., Panwar, A., Mohan, J., Kwatra, N., Gulavani, B.~S., Tumanov, A., and Ramjee, R.}
\newblock Taming throughout-latency tradeoff in llm inference with sarathi-serve.
\newblock In {\em OSDI 2024\/} (May 2024).

\bibitem{clipper}
{\sc Crankshaw, D., Wang, X., Zhou, G., Franklin, M.~J., Gonzalez, J.~E., and Stoica, I.}
\newblock Clipper: A {Low-Latency} online prediction serving system.
\newblock In {\em 14th USENIX Symposium on Networked Systems Design and Implementation (NSDI 17)\/} (Boston, MA, Mar. 2017), USENIX Association, pp.~613--627.

\bibitem{envoy}
{\sc {Envoy Authors}}.
\newblock {\em Envoy Documentation}, 2024.
\newblock \url{https://www.envoyproxy.io/docs}, accessed on 16 Sep 2024.

\bibitem{envoy-lb}
{\sc {Envoy Authors}}.
\newblock {\em Envoy Load Balancer Documentation}, 2024.
\newblock \url{https://www.envoyproxy.io/docs/envoy/latest/intro/arch_overview/upstream/load_balancing/load_balancers}, accessed on 17 Sep 2024.

\bibitem{ltr}
{\sc Fu, Y., Zhu, S., Su, R., Qiao, A., Stoica, I., and Zhang, H.}
\newblock Efficient llm scheduling by learning to rank, 2024.

\bibitem{lora}
{\sc Hu, E.~J., yelong shen, Wallis, P., Allen-Zhu, Z., Li, Y., Wang, S., Wang, L., and Chen, W.}
\newblock Lo{RA}: Low-rank adaptation of large language models.
\newblock In {\em International Conference on Learning Representations\/} (2022).

\bibitem{istio}
{\sc {Istio Authors}}.
\newblock {\em Istio Documentation}, 2024.
\newblock \url{https://istio.io/latest/docs/}, accessed on 16 Sep 2024.

\bibitem{istio-lb}
{\sc {Istio Authors}}.
\newblock {\em Istio Load Balancer Documentation}, 2024.
\newblock \url{https://istio.io/latest/docs/concepts/traffic-management/#load-balancing-options}, accessed on 17 Sep 2024.

\bibitem{s3}
{\sc Jin, Y., Wu, C.-F., Brooks, D., and Wei, G.-Y.}
\newblock S3: Increasing {GPU} utilization during generative inference for higher throughput.
\newblock In {\em Thirty-seventh Conference on Neural Information Processing Systems\/} (2023).

\bibitem{knative}
{\sc {KNative Authors}}.
\newblock {\em KNative Documentation}, 2024.
\newblock \url{https://knative.dev/docs/}, accessed on 16 Sep 2024.

\bibitem{knative-lb}
{\sc {KNative Authors}}.
\newblock {\em KNative Load Balancer Documentation}, 2024.
\newblock \url{https://knative.dev/docs/serving/load-balancing/}, accessed on 17 Sep 2024.

\bibitem{dynamoe}
{\sc Kossmann, F., Jia, Z., and Aiken, A.}
\newblock Optimizing mixture of experts using dynamic recompilations, 2022.

\bibitem{vetl}
{\sc Kossmann, F., Wu, Z., Lai, E., Tatbul, N., Cao, L., Kraska, T., and Madden, S.}
\newblock Extract-transform-load for video streams.
\newblock {\em Proc. VLDB Endow. 16}, 9 (May 2023), 2302–2315.

\bibitem{cascadeserve}
{\sc Kossmann, F., Wu, Z., Turk, A., Tatbul, N., Cao, L., and Madden, S.}
\newblock Cascadeserve: Unlocking model cascades for inference serving, 2024.

\bibitem{kserve}
{\sc {KServe Authors}}.
\newblock {\em KServe Documentation}, 2024.
\newblock \url{https://kserve.github.io/website/latest/}, accessed on 16 Sep 2024.

\bibitem{kubernetes}
{\sc {Kubernetes Authors}}.
\newblock {\em Kubernetes Documentation}, 2024.
\newblock \url{https://kubernetes.io/docs/home/}, accessed on 16 Sep 2024.

\bibitem{vllm}
{\sc Kwon, W., Li, Z., Zhuang, S., Sheng, Y., Zheng, L., Yu, C.~H., Gonzalez, J., Zhang, H., and Stoica, I.}
\newblock Efficient memory management for large language model serving with pagedattention.
\newblock In {\em Proceedings of the 29th Symposium on Operating Systems Principles\/} (New York, NY, USA, 2023), SOSP '23, Association for Computing Machinery, p.~611–626.

\bibitem{alizadeh}
{\sc Mao, H., Schwarzkopf, M., Venkatakrishnan, S.~B., Meng, Z., and Alizadeh, M.}
\newblock Learning scheduling algorithms for data processing clusters.
\newblock In {\em Proceedings of the ACM Special Interest Group on Data Communication\/} (New York, NY, USA, 2019), SIGCOMM '19, Association for Computing Machinery, p.~270–288.

\bibitem{mlflow}
{\sc {MLFlow Authors}}.
\newblock {\em MLFlow Documentation}, 2024.
\newblock \url{https://mlflow.org/docs/latest/index.html}, accessed on 16 Sep 2024.

\bibitem{ray}
{\sc Moritz, P., Nishihara, R., Wang, S., Tumanov, A., Liaw, R., Liang, E., Elibol, M., Yang, Z., Paul, W., Jordan, M.~I., and Stoica, I.}
\newblock Ray: A distributed framework for emerging {AI} applications.
\newblock In {\em 13th USENIX Symposium on Operating Systems Design and Implementation (OSDI 18)\/} (Carlsbad, CA, Oct. 2018), USENIX Association, pp.~561--577.

\bibitem{splitwise}
{\sc Patel, P., Choukse, E., Zhang, C., Shah, A., Goiri, I., Maleki, S., and Bianchini, R.}
\newblock Splitwise: Efficient generative llm inference using phase splitting.
\newblock In {\em ISCA\/} (June 2024).

\bibitem{deepspeed}
{\sc Rasley, J., Rajbhandari, S., Ruwase, O., and He, Y.}
\newblock Deepspeed: System optimizations enable training deep learning models with over 100 billion parameters.
\newblock In {\em Proceedings of the 26th ACM SIGKDD International Conference on Knowledge Discovery \& Data Mining\/} (New York, NY, USA, 2020), KDD '20, Association for Computing Machinery, p.~3505–3506.

\bibitem{infaas}
{\sc Romero, F., Li, Q., Yadwadkar, N.~J., and Kozyrakis, C.}
\newblock {INFaaS}: Automated model-less inference serving.
\newblock In {\em 2021 USENIX Annual Technical Conference (USENIX ATC 21)\/} (July 2021), USENIX Association, pp.~397--411.

\bibitem{distilbert}
{\sc Sanh, V., Debut, L., Chaumond, J., and Wolf, T.}
\newblock Distilbert, a distilled version of bert: smaller, faster, cheaper and lighter, 2020.

\bibitem{trail}
{\sc Shahout, R., Malach, E., Liu, C., Jiang, W., Yu, M., and Mitzenmacher, M.}
\newblock Don't stop me now: Embedding based scheduling for llms, 2024.

\bibitem{sprpt}
{\sc Shahout, R., and Mitzenmacher, M.}
\newblock Skippredict: When to invest in predictions for scheduling, 2024.

\bibitem{workload}
{\sc Shahrad, M., Fonseca, R., Goiri, I., Chaudhry, G., Batum, P., Cooke, J., Laureano, E., Tresness, C., Russinovich, M., and Bianchini, R.}
\newblock Serverless in the wild: Characterizing and optimizing the serverless workload at a large cloud provider.
\newblock In {\em 2020 USENIX Annual Technical Conference (USENIX ATC 20)\/} (July 2020), USENIX Association, pp.~205--218.

\bibitem{slora1}
{\sc Sheng, Y., Cao, S., Li, D., Hooper, C., Lee, N., Yang, S., Chou, C., Zhu, B., Zheng, L., Keutzer, K., Gonzalez, J.~E., and Stoica, I.}
\newblock S-lora: Serving thousands of concurrent lora adapters.
\newblock {\em arXiv preprint arXiv:2311.03285\/} (2023).

\bibitem{slora2}
{\sc Sheng, Y., Cao, S., Li, D., Zhu, B., Li, Z., Zhuo, D., Gonzalez, J.~E., and Stoica, I.}
\newblock Fairness in serving large language models.
\newblock {\em arXiv preprint arXiv:2401.00588\/} (2023).

\bibitem{flexgen}
{\sc Sheng, Y., Zheng, L., Yuan, B., Li, Z., Ryabinin, M., Chen, B., Liang, P., R\'{e}, C., Stoica, I., and Zhang, C.}
\newblock Flexgen: high-throughput generative inference of large language models with a single gpu.
\newblock In {\em Proceedings of the 40th International Conference on Machine Learning\/} (2023), ICML'23, JMLR.org.

\bibitem{llumnix}
{\sc Sun, B., Huang, Z., Zhao, H., Xiao, W., Zhang, X., Li, Y., and Lin, W.}
\newblock Llumnix: Dynamic scheduling for large language model serving.
\newblock In {\em 18th USENIX Symposium on Operating Systems Design and Implementation (OSDI 24)\/} (Santa Clara, CA, July 2024), USENIX Association, pp.~173--191.

\bibitem{trt-llm}
{\sc {TensorRT-LLM Team}}.
\newblock {\em {TensorRT-LLM Repository}}, 2024.
\newblock \url{https://github.com/NVIDIA/TensorRT-LLM}, accessed on 17 Sep 2024.

\bibitem{burstgpt}
{\sc Wang, Y., Chen, Y., Li, Z., Kang, X., Tang, Z., He, X., Guo, R., Wang, X., Wang, Q., Zhou, A.~C., and Chu, X.}
\newblock Burstgpt: A real-world workload dataset to optimize llm serving systems, 2024.

\bibitem{fastserve}
{\sc Wu, B., Zhong, Y., Zhang, Z., Huang, G., Liu, X., and Jin, X.}
\newblock Fast distributed inference serving for large language models, 2023.

\bibitem{stage}
{\sc Wu, Z., Marcus, R., Liu, Z., Negi, P., Nathan, V., Pfeil, P., Saxena, G., Rahman, M., Narayanaswamy, B., and Kraska, T.}
\newblock Stage: Query execution time prediction in amazon redshift.
\newblock In {\em Companion of the 2024 International Conference on Management of Data\/} (New York, NY, USA, 2024), SIGMOD/PODS '24, Association for Computing Machinery, p.~280–294.

\bibitem{orca}
{\sc Yu, G.-I., Jeong, J.~S., Kim, G.-W., Kim, S., and Chun, B.-G.}
\newblock Orca: A distributed serving system for {Transformer-Based} generative models.
\newblock In {\em 16th USENIX Symposium on Operating Systems Design and Implementation (OSDI 22)\/} (Carlsbad, CA, July 2022), USENIX Association, pp.~521--538.

\bibitem{brad}
{\sc Yu, G.~X., Wu, Z., Kossmann, F., Li, T., Markakis, M., Ngom, A., Madden, S., and Kraska, T.}
\newblock Blueprinting the cloud: Unifying and automatically optimizing cloud data infrastructures with brad.
\newblock {\em Proc. VLDB Endow. 17}, 11 (Aug. 2024), 3629–3643.

\bibitem{shepherd}
{\sc Zhang, H., Tang, Y., Khandelwal, A., and Stoica, I.}
\newblock {SHEPHERD}: Serving {DNNs} in the wild.
\newblock In {\em 20th USENIX Symposium on Networked Systems Design and Implementation (NSDI 23)\/} (Boston, MA, Apr. 2023), USENIX Association, pp.~787--808.

\bibitem{sglang}
{\sc Zheng, L., Yin, L., Xie, Z., Sun, C., Huang, J., Yu, C.~H., Cao, S., Kozyrakis, C., Stoica, I., Gonzalez, J.~E., Barrett, C., and Sheng, Y.}
\newblock Sglang: Efficient execution of structured language model programs, 2024.

\bibitem{pia}
{\sc Zheng, Z., Ren, X., Xue, F., Luo, Y., Jiang, X., and You, Y.}
\newblock Response length perception and sequence scheduling: An llm-empowered llm inference pipeline.
\newblock In {\em Advances in Neural Information Processing Systems\/} (2023), A.~Oh, T.~Naumann, A.~Globerson, K.~Saenko, M.~Hardt, and S.~Levine, Eds., vol.~36, Curran Associates, Inc., pp.~65517--65530.

\end{thebibliography}



\end{document}